\documentclass[letterpaper, 10 pt, conference]{ieeeconf}  

\IEEEoverridecommandlockouts                              

\usepackage{graphics} 
\usepackage{epsfig} 
\usepackage{times} 
\usepackage{amsmath} 
\usepackage{amssymb}  

\usepackage{booktabs, multirow} 
\usepackage{float}

\usepackage{algorithm}
\usepackage{algpseudocode}
\usepackage[table]{xcolor}

\usepackage{caption}
\usepackage{subcaption}
\usepackage{pifont}
\newcommand{\cmark}{\ding{51}}%
\newcommand{\xmark}{\ding{55}}%

\usepackage{xcolor}
\usepackage{tablefootnote}
\newcommand{\gb}[1]{\cellcolor{gray!50}\textbf{#1}}

\title{\LARGE \bf

FalconTrack: Photorealistic Auto-Labeled Perception and\\Physics-Aware Vision-Based Aerial Tracking
}

\author{Yan Miao$^{1}$, Karteek Gandiboyina$^{1}$, Noah Giles$^{1}$, Hideki Okamoto$^{2}$,
\\ 
Bardh Hoxha$^{2}$, Georgios Fainekos$^{2}$ and Sayan Mitra$^{1}$
\thanks{{\footnotesize$^{1}$ Yan Miao, Karteek Gandiboyina, Noah Giles and Sayan Mitra are with the Department of Electrical and Computer Engineering, University of Illinois at Urbana Champaign {\tt\footnotesize \{yanmiao2, mkg7, negiles2, mitras\}@illinois.edu}}}
\thanks{{\footnotesize$^{2}$ Hideki Okamoto, Bardh Hoxha and Georgios Fainekos are with Toyota Motor North America R\&D {\tt\footnotesize \{firstname.lastname\}@toyota.com}}}
}

\begin{document}
\bstctlcite{BSTcontrol}

\raggedbottom

\maketitle
\thispagestyle{empty}
\pagestyle{empty}

\begin{abstract}
Vision-based aerial tracking is critical in GPS-denied environments.
Reliable perception for tracking depends on large-scale labeled data, yet most photorealistic datasets rely on heavy manual annotation and are time-consuming to produce.
We present \textit{FalconTrack}, a unified perception-and-tracking framework that (i) leverages a photorealistic editable simulator for automated label generation and (ii) combines multi-head perception with physics-aware tracking for zero-shot sim-to-real transfer.
FalconTrack provides an automated labeling pipeline in a Gaussian Splatting simulator that isolates target Gaussians from short object videos and composites them with randomized backgrounds to generate RGB, mask, class, and 6-DoF pose labels, producing about 10k labeled images in under 20 minutes.
Using this dataset, we train a multi-head perception module with staged learning and reprojection consistency, and fuse its outputs with class-conditioned dynamics priors in an EKF for tracking.
Our perception model outperforms two baselines and reaches 96--100\% class accuracy in zero-shot sim-to-real transfer on three geometrically diverse objects and two environments, while maintaining consistent performance in unseen simulated and real scenes.
In real hardware closed-loop visual tracking, the onboard system runs at about 25\,Hz and achieves 100\% success in sim-to-real F1-tenth and gate tracking in five trajectories across two environments, while a mask-centered vision baseline drops to 60\% success on F1-tenth during fast out-of-view scenarios.
\end{abstract}

\section{Introduction}

Autonomous vision-based aerial tracking is important for applications such as transportation, environmental monitoring, and search and rescue, especially in GPS-denied environments where ground-truth state is unavailable.

\begin{figure}[tbp]
    \centering
    \includegraphics[width=\linewidth]{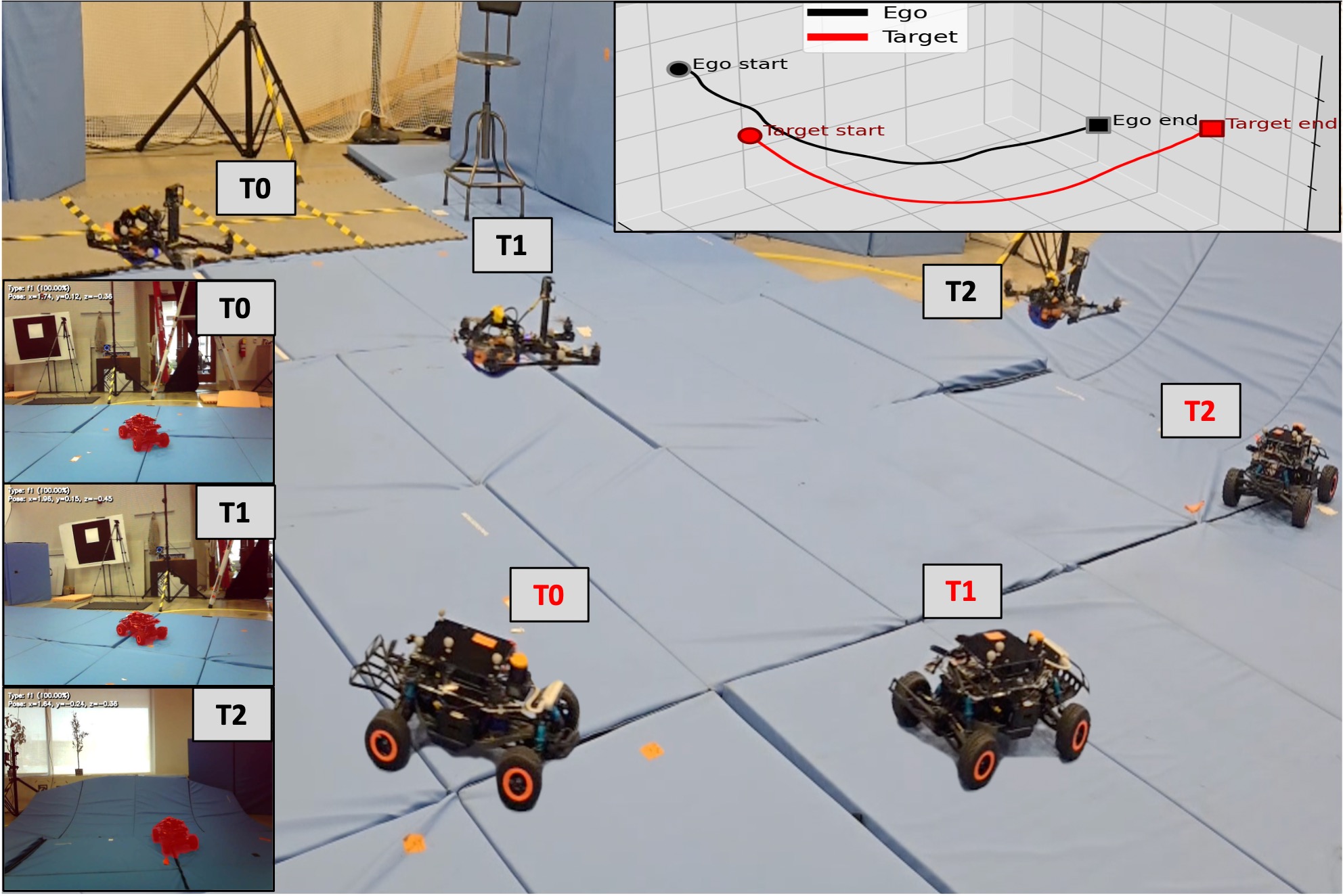}
    \caption{\small{\textbf{FalconTrack for Real-World Tracking:}
    We show a time-lapse of an airborne ego quadrotor tracking a ground F1-tenth \cite{pmlr-v123-o-kelly20a} target vehicle during a left-turn trajectory using vision. Time labels (T0--T2) use black text for ego snapshots and red text for target snapshots; the left insets show onboard RGB images with predicted masks and pose estimates at T0--T2, and the top-right plot visualizes the 3D trajectories of the ego (black) and target (red).}}
    \label{fig:tracking-demo}
\end{figure}

\begin{figure*}[htbp]
    \centering
    \includegraphics[width=0.9\linewidth]{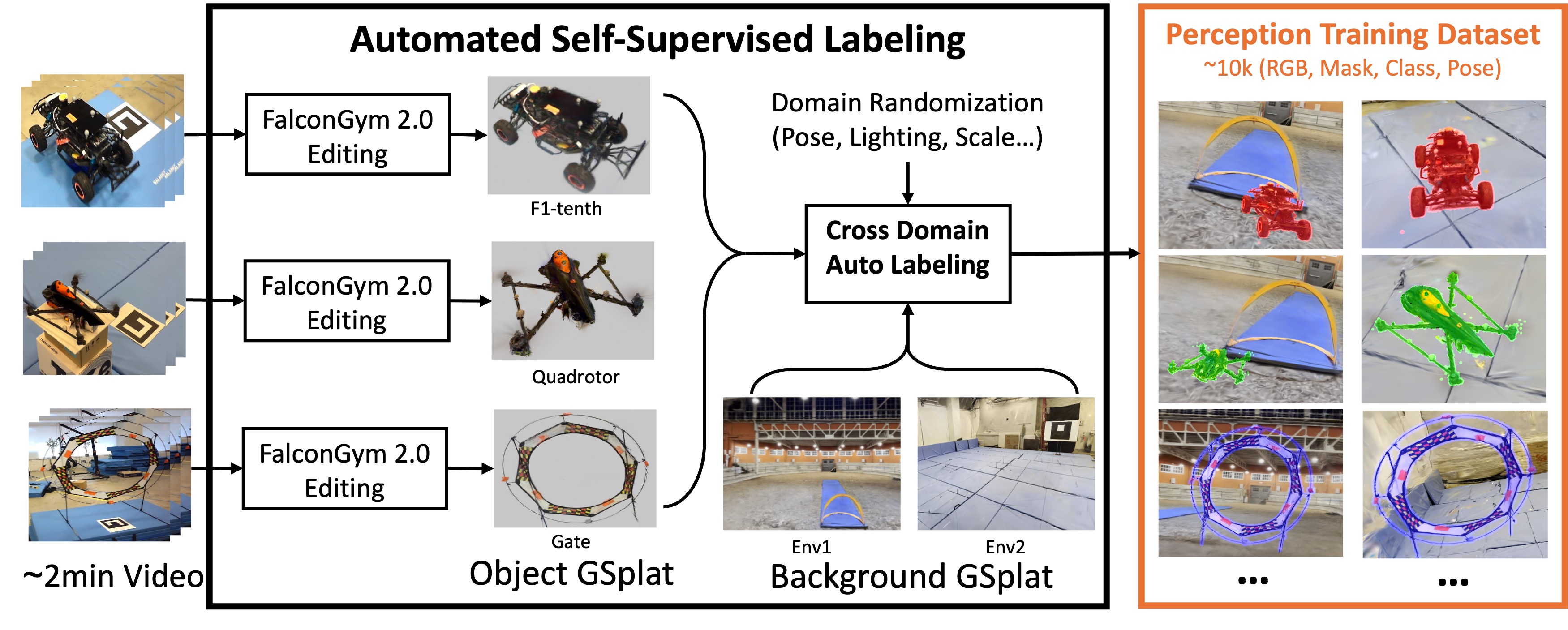}
    \caption{\small{
    \textbf{Auto-Labeling Pipeline for a Photorealistic Perception Dataset:}
    From a 2-minute real video per object, we construct object GSplats and combine them with diverse background GSplats and domain randomization using the FalconGym 2.0 Edit API \cite{MiaoEtAl:ICRA26}. Using known camera matrices and rendering transforms, we automatically compute pixel-accurate masks and 6-DoF poses for each frame, yielding a large-scale, diverse training dataset of RGB images, masks, class labels, and 6-DoF poses without manual annotation.
    }}
    \label{fig:auto-label}
\end{figure*}

Reliable vision-based tracking typically relies on accurate perception models trained on large-scale labeled datasets.
However, many existing perception datasets rely heavily on human annotation, which is costly, time-consuming, potentially error-prone, and difficult to scale to new irregular objects (non-analytic shapes or non-parametric geometry).
For example, ImageNet \cite{5206848}, widely used to train object detectors and classifiers such as YOLO \cite{7780460}, required thousands of Amazon Mechanical Turk workers \cite{10.1145/3351095.3375709} to manually label categories; BDD100K \cite{bdd100k}, widely used for self-driving policy training, also relies on human annotation for 100K images with bounding boxes and lane markings.
Some simulators (e.g., CARLA \cite{Dosovitskiy17} and Gazebo \cite{1389727}) can auto-generate labels. For example, KITTI-CARLA \cite{deschaud2021kitticarla} automatically provides labels for lidar, instance segmentation, and 3D object bounding boxes in CARLA. 
However, images from such simulators often lack photorealism, causing a sim-to-real gap when trained perception is deployed in the real world.
Diffusion-based generative models such as GLIDE \cite{Nichol2021GLIDETP} can synthesize photorealistic 2D images from text, but they do not guarantee that generated objects match the target appearance and more importantly do not provide 6-DoF pose labels.
Recent advances in photorealistic simulators based on 3D Gaussian Splatting \cite{kerbl3Dgaussians}, such as FalconGym 2.0 \cite{MiaoEtAl:ICRA26}, show promise by training a gate mask detector in simulation and deploying zero-shot in the real world. 
However, their gate labeling remains semi-manual, relying on measured gate size and analytic geometry (e.g., a ring defined in a canonical frame), and is demonstrated on a single gate object rather than irregular targets (e.g., an F1-tenth vehicle or a quadrotor).
To bridge the gap between photorealistic image synthesis and automated labeling of irregular objects, we propose an automated labeling pipeline.
From a 2-minute video, we reconstruct a 3D Gaussian Splatting scene, use SAGA \cite{cen2023saga} to isolate target-related splats, and fuse them with diverse background GSplats under domain randomization to improve generalization and reduce sim-to-real gaps.
Given the camera intrinsics and extrinsics used for rendering, we project the 3D Gaussians into the image plane to generate pixel-accurate masks and compute the 6-DoF target pose directly from the known rendering transform.
This yields a photorealistic and auto-labeled dataset for perception of irregular objects without any manual annotation.

\begin{figure}[htbp]
    \centering
    \includegraphics[width=\linewidth]{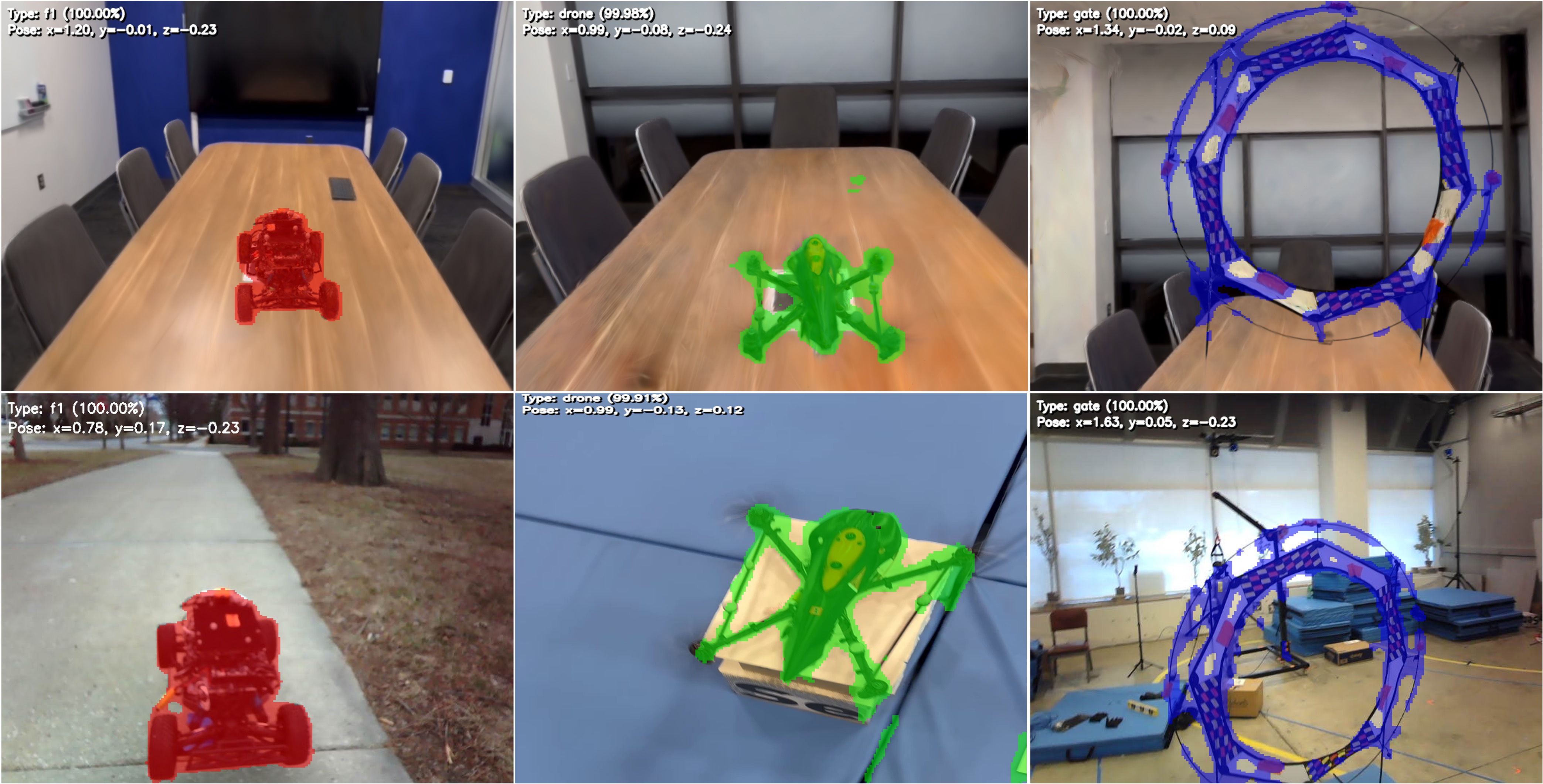}
    \caption{\small{
        \textbf{Qualitative Perception in Unseen GSplats and Real World:} The first row shows mask and 6-DoF pose estimations on RGB images rendered by FalconGym 2.0 with the target object in unseen background GSplats. The second row shows perception on targets in real-world unseen environments. Mask colors indicate predicted class: red (F1-tenth), green (quadrotor), and blue (gate).
    }}
    \label{fig:perception-unknown-background}
\end{figure}

A well-labeled perception dataset can provide the basis for robust closed-loop vision-based aerial tracking across diverse target types and environments. 
Drone racing is a representative special case of tracking a sequence of static gates. 
Prior racing work \cite{DBLP:conf/rss/GelesBRX024} \cite{kaufmann2023champion} trains a gate detector and connects it to downstream neural control; although it can reach high speed (e.g., 11\,m/s), it does not generalize to new tracks or object types.
FalconGym 2.0 \cite{MiaoEtAl:ICRA26} uses a single policy to traverse different tracks but still focuses on one object type (gate). 
A transformer-based unified perception module \cite{10610111} generalizes tracking across target types and uses a model-free controller to follow the detected 2D bounding-box center, but it does not exploit target-specific dynamics and 6 DoF pose estimates.
To address these limitations, we move beyond per-object 2D detection and design a perception module that predicts target class and full 6-DoF pose from our auto-labeled data, trained with staged learning and a reprojection-consistency loss. 
Building on this 6-DoF perception, we perform closed-loop tracking of three target objects in simulation and real experiments, using class-conditioned dynamics priors to smooth pose estimates and recover from temporary out-of-view events.
Experiments show that our unified perception module classifies three irregular objects with 100\% accuracy in simulation and reaches 96--100\% class accuracy when deployed zero-shot to the real world. 
Qualitative results in Figure \ref{fig:perception-unknown-background} also show consistent perception in unseen simulated and real environments.
In vision-based closed-loop tracking, our physics-aware tracking controller achieves 100\% success across all simulation trajectories and 100\% success in real F1-tenth and gate tracking, while a vision-only baseline shows lower success (60\% on F1-tenth, 100\% on gate) and larger tracking errors during fast out-of-view scenarios. 

In summary, we introduce \textit{FalconTrack}, a unified perception and tracking framework for vision-based aerial tracking. Our contributions are threefold:
(1) an automated labeling pipeline that generates photorealistic labels for diverse irregular targets without manual annotation;
(2) a unified multi-head perception with staged training and reprojection consistency that zero-shot sim-to-real transfers on three geometrically diverse objects and backgrounds; 
(3) a physics-aware tracking controller with class-conditioned dynamics priors that enables closed-loop visual tracking on a real quadrotor across multiple target types, trajectories, and environments.


\section{Related Work}

\paragraph{Labeled Perception Datasets}
Large-scale perception datasets such as ImageNet \cite{5206848}, BDD100K \cite{bdd100k}, and Ego-Exo4D \cite{Grauman2023EgoExo4DUS} provide category labels and bounding boxes, but they rely heavily on manual annotation and do not scale well to new irregular objects. 
Synthetic data from simulators like CARLA \cite{Dosovitskiy17} and Gazebo \cite{1389727} can be automatically labeled but often suffer from limited photorealism, which hinders sim-to-real transfer. 
The photorealistic Gaussian Splatting-based FalconGym 2.0 \cite{MiaoEtAl:ICRA26} improves realism, yet its semi-manual labeling assumes analytic object geometry and could not scale to irregular objects.

\paragraph{Perception in Computer Vision}
Object-centric perception is widely used in robotics. 
CNN-based instance segmentation models such as Mask R-CNN \cite{8237584} provide high-quality masks, while real-time approaches such as YOLACT \cite{10.1109/TPAMI.2020.3014297} trade some accuracy for speed; however, both depend on large manually labeled datasets and do not estimate pose directly. 
Learning-based 6D estimators have also been used in manipulation and tabletop robotics.
PoseCNN \cite{xiang2018posecnn} and PVNet \cite{peng2019pvnet} handle clutter and partial occlusion; RGB-D pipelines like DenseFusion \cite{wang2019densefusion} improve accuracy with depth but add sensing cost. 
%

\paragraph{Vision-based Aerial Tracking}
Vision-based aerial racing systems demonstrate high-speed gate tracking with onboard perception and control \cite{kaufmann2023champion,DBLP:conf/rss/GelesBRX024}. 
These methods achieve impressive, human-champion-level performance, but they are typically trained for customized tracks and overfit to a single target type. 
Related work in aerial navigation \cite{11247178,miao2025falconwingultralightindoorfixedwing} and tracking \cite{bosello2026oyo} focuses on structured gates and environments, with limited generalization to irregular objects or unseen backgrounds.

\paragraph{Photorealistic Sim-to-Real Transfer in Robotics}
Neural scene representations such as NeRF \cite{10.1145/3503250} and 3D Gaussian Splatting \cite{kerbl3Dgaussians} enable photorealistic simulators, motivating sim-to-real applications in manipulation \cite{robosplat} and self-driving \cite{yan2024street}. 
For visual aerial robotics, Sous-Vide \cite{low2024sousvide}, Grad-Nav \cite{chen2025gradnavefficientlylearningvisual}, FalconGym \cite{11247178} and FalconWing \cite{miao2025falconwingultralightindoorfixedwing} report strong sim-to-real transfer. 
However, these approaches often assume identical one-to-one digital twins between simulation and reality. 
FalconGym 2.0 \cite{MiaoEtAl:ICRA26} broadens track diversity but remains focused on a single static object type (gates).
\section{Methodology}

\begin{figure}[htbp]
    \centering
    \captionsetup[subfigure]{labelfont=bf}
    \begin{subfigure}[b]{0.49\linewidth}
        \centering
        \includegraphics[width=\textwidth]{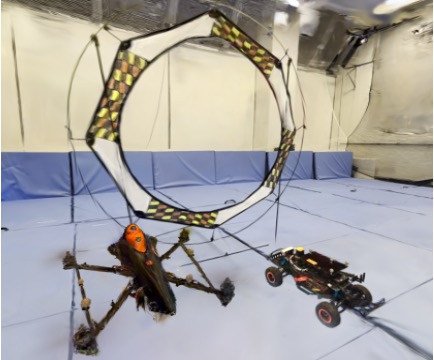}
        \caption{\textbf{FalconGym 2.0 Simulation}}
        \label{fig:falcongym-setup}
    \end{subfigure}
    \hfill 
    \begin{subfigure}[b]{0.49\linewidth}
        \centering
        \includegraphics[width=\textwidth]{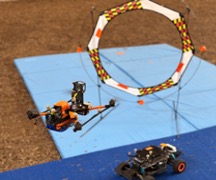}
        \caption{\textbf{Real World Setup}}
        \label{fig:hardware-setup}
    \end{subfigure}
    
    
    \begin{subfigure}[b]{\linewidth}
        \centering
        \includegraphics[width=\textwidth]{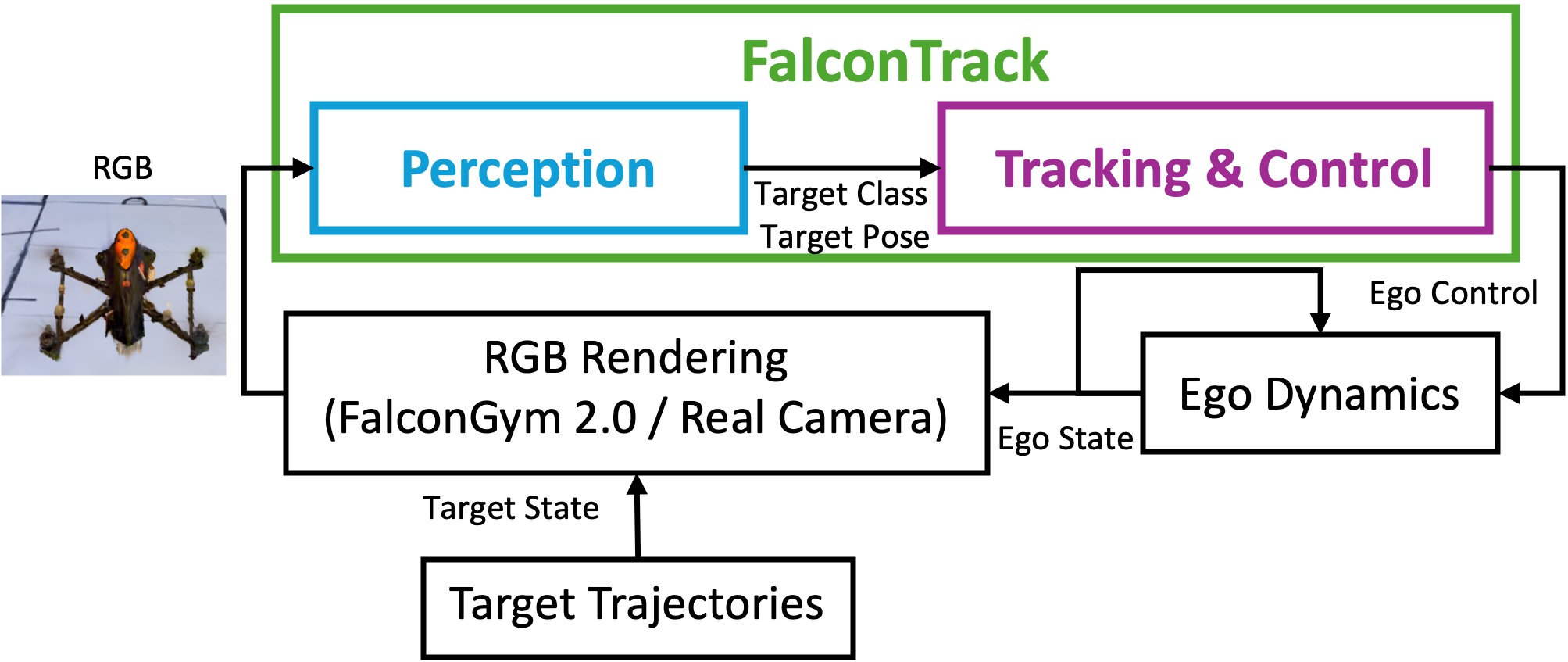}
        \caption{\textbf{Ego Closed-loop for Vision-based Aerial Tracking}}
        \label{fig:close-loop}
    \end{subfigure}
    \caption{FalconTrack (perception and tracking \& control modules) is trained in FalconGym 2.0 and then deployed zero-shot from simulation to real-world target tracking.}
    
\end{figure}

In this paper, we address vision-based aerial tracking of dynamically moving targets using an onboard RGB camera on a quadrotor.
We first develop FalconTrack (green box in Figure \ref{fig:close-loop}) in the photorealistic FalconGym 2.0 simulation environment \cite{MiaoEtAl:ICRA26}, and then deploy it zero-shot from simulation to a real quadrotor for tracking across different target classes, trajectories, and background environments. 
One real-world tracking scenario is shown in Figure \ref{fig:tracking-demo}.

FalconTrack has two components: (i) a unified perception module (Section \ref{sec:perception}) that predicts target class, segmentation mask, and relative 6-DoF target pose, trained with automatically generated labels using a staged training algorithm and reprojection consistency;
and (ii) a physics-aware tracking and control module (Section \ref{sec:tracking-controller}) that fuses perception with target dynamics priors to follow the target.
In this paper, each run assumes tracking a single target, and target switching within a run is not considered.

\subsection{Unified Perception Module}
\label{sec:perception}

As shown in Figure \ref{fig:close-loop}, the FalconTrack framework is modular, separating perception from tracking and control. 
This design simplifies debugging, improves interpretability, and enables reuse of the same perception module across downstream control tasks. 
We develop a perception module that takes an RGB image as input and outputs the target class, mask, and 6-DoF pose in the camera frame (treated as the ego's body frame in this paper).

\subsubsection{Automated Labeling Dataset}
As previously discussed in the Introduction and Related Work, obtaining large-scale photorealistic data with accurate 6-DoF supervision remains a challenge.
Existing large datasets often require costly and error-prone manual annotation, while traditional synthetic pipelines provide auto-labeling but introduce a larger sim-to-real gap due to limited photorealism.
To address this gap, we build an automated labeling pipeline that renders photorealistic RGB images while providing precise ground-truth class, mask, and pose labels without human annotation.

As shown in Figure \ref{fig:auto-label}, we first collect a short handheld video of each target (about 2 minutes, $\sim$200 frames) with multiple viewpoints and minimal motion blur using the onboard camera. 
We then reconstruct a Gaussian Splatting (GSplat) model of the target at real-world scale using a calibrated marker.
Next, we apply SAGA \cite{cen2023saga} and FalconGym 2.0's Edit API \cite{MiaoEtAl:ICRA26} to separate the object GSplat from the background (see \cite{MiaoEtAl:ICRA26} for calibration and Gaussian Splat editing details). 
To diversify appearances, we fuse the object GSplat with multiple background GSplats from FalconGym 2.0 \cite{MiaoEtAl:ICRA26} using the same Edit API, then render RGB images with domain randomization over pose, scale, and lighting to improve diversity and potentially reduce sim-to-real gap.
Finally, since we have full access to ground truth, including object GSplat pose, scale, color, and camera intrinsic/extrinsic parameters, we compute the target's relative 6-DoF pose in the camera frame via geometric calibration and project the segmented target Gaussians to generate pixel-accurate masks.

With roughly 20 minutes of automated processing (10 minutes for GSplat reconstruction and editing, 10 minutes for mask labeling and pose computation) on a desktop RTX 4090 GPU, we generate $\sim$10k labeled samples (RGB, class, mask, pose) from three objects across two environments, as shown in Figure \ref{fig:auto-label}. 
We also mix in RGB images without any targets to take into consideration of empty scenes and temporary target out-of-view events. 
This automated pipeline produces fast and accurate labels at scale and provides diverse training data for our unified perception module.

\subsubsection{Perception Architecture}

\begin{figure}[htbp]
    \centering
    \includegraphics[width=\linewidth]{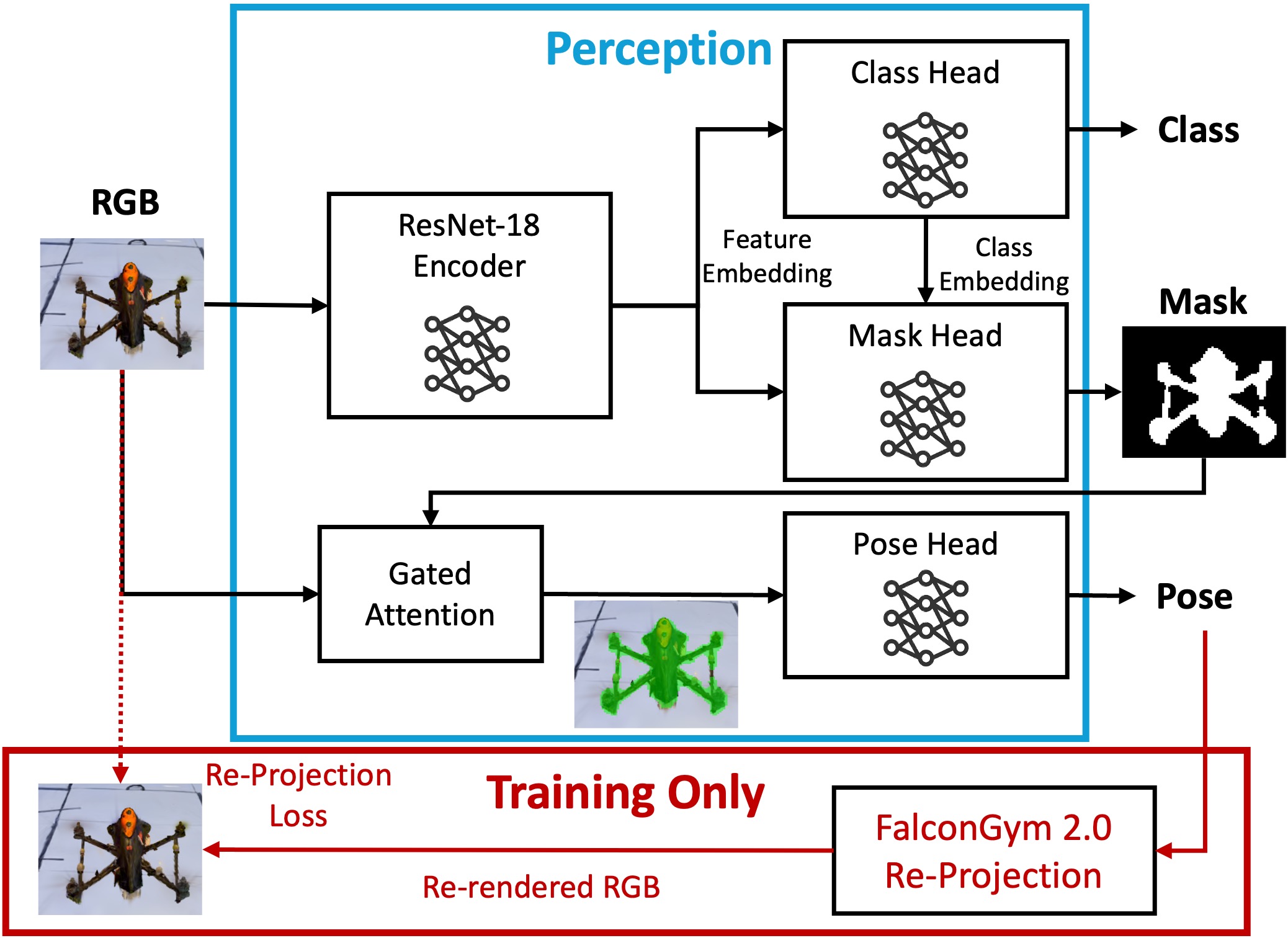}
    \caption{\small{
    \textbf{Perception Architecture:}
    a shared encoder extracts features that branch into a classification head and a mask head.
    The predicted mask conditions a gated-attention pose head to regress the relative 6-DoF target pose.
    During training, we use FalconGym 2.0 to re-render targets at predicted poses and apply reprojection loss to improve geometric consistency and pose estimation.
    }}
    \label{fig:perception-module}
\end{figure}

With the cross-domain auto-labeled dataset, we train a unified perception module that generalizes across multiple objects and backgrounds while remaining practical for onboard deployment. 
Our design follows two goals: (1) maintaining a compact model for real-time inference on the quadrotor, and (2) enforcing geometric consistency using FalconGym 2.0's differentiable re-rendering capability. 
As shown in Figure \ref{fig:perception-module}, the network uses a shared ResNet-18 encoder with three heads: classification, segmentation, and pose. 
This shared backbone reduces parameters and encourages a common feature representation across tasks, improving both efficiency and generalization. 
The classification head predicts the target class (e.g., F1-tenth \cite{pmlr-v123-o-kelly20a}, quadrotor, gate), which the tracking controller uses to select class-specific dynamics priors (Section \ref{sec:tracking-controller}); these class features also support the mask head. 
Prior work has explored both RGB-only transformer-based pose regression and mask-only pose estimation: 
RGB-only ViT models can perform well in aerial navigation \cite{11247178}, but their larger size and tendency to overfit background appearance make high-rate onboard closed-loop deployment challenging across environments; 
mask-only approaches can be effective for static-gate tasks \cite{DBLP:conf/rss/GelesBRX024}, but masks alone can be ambiguous for symmetric objects (e.g., yaw 0 vs. $\pi$ for a quadrotor). 
Motivated by these limitations, we regress 6-DoF pose by combining RGB and foreground-mask features through gated attention, which suppresses background clutter while preserving appearance cues needed for orientation. 
Finally, we include a reprojection term in the loss function: the target is re-rendered in FalconGym 2.0 at the predicted pose and compared with the input RGB using SSIM \cite{1284395}. 
This loss improves geometric consistency and corrects visually significant small pose errors, especially in yaw. 

The resulting multi-head perception module has 15M parameters and runs at about 30\,Hz on an NVIDIA Jetson Orin (the quadrotor's onboard computer).

\subsubsection{Perception Training}

\begin{algorithm}[htbp]
    \caption{\small{Staged Training Algorithm for Perception Module}}
    \label{alg:perception-training}
    \begin{algorithmic}[1]
    \Require $\mathcal{D}=\{(I_i, m_i, y_i, p_i)\}_i^N$: RGB, mask, class, 6D pose.
    \Ensure Perception module $\pi_\theta = (\pi_{\mathrm{class}}, \pi_{\mathrm{mask}}, \pi_{\mathrm{pose}})$.
    \State \textbf{Stage 1:} Freeze $\pi_{\mathrm{mask}}$ and $\pi_{\mathrm{pose}}$; train $\pi_{\mathrm{class}}$ only.
    \State \textbf{Stage 2:} Freeze $\pi_{\mathrm{pose}}$; train $\pi_{\mathrm{mask}}$ and $\pi_{\mathrm{class}}$.
    \State \textbf{Stage 3:} Train full $\pi_\theta: I \mapsto (\hat{m}, \hat{y}, \hat{p})$
    \For{each batch in Stage 3}
      \State $\mathcal{L} \gets \lambda_1 \cdot \mathcal{L}_{\mathrm{class}} + \lambda_2 \cdot \mathcal{L}_{\mathrm{mask}} + \lambda_3 \cdot \mathcal{L}_{\mathrm{pose}}$
      
      \State $\mathcal{L} \gets \mathcal{L} + \lambda_4 \cdot \mathrm{SSIM}(\mathrm{Render}(\hat{p}), I)$ \Comment{\small{Reproject Loss}}
    
      \State Gradient Descend with $\mathcal{L}$ to update $\theta$
    \EndFor
    \end{algorithmic}
\end{algorithm}

We train the multi-head network with a staged curriculum (Algorithm \ref{alg:perception-training}) rather than end-to-end from scratch. 
Since the three heads are sequentially coupled, errors in early predictions can destabilize later heads (e.g., poor masks corrupt pose estimation). 
To improve stability and convergence, we train the heads sequentially using the staged procedure. 
In Stage 1, we train only the class head for 20 epochs while freezing the mask and pose heads. 
In Stage 2, we train the class and mask heads for 30 epochs while keeping the pose head frozen. 
In Stage 3, we train all heads jointly for 200 epochs with reprojection loss enabled (only when a target is present; skipped for no-object samples). 
Training takes about 3 hours on a 4090 GPU.

\subsection{Tracking Controller}
\label{sec:tracking-controller}

\begin{figure}[htbp]
    \centering
    \includegraphics[width=0.9\linewidth]{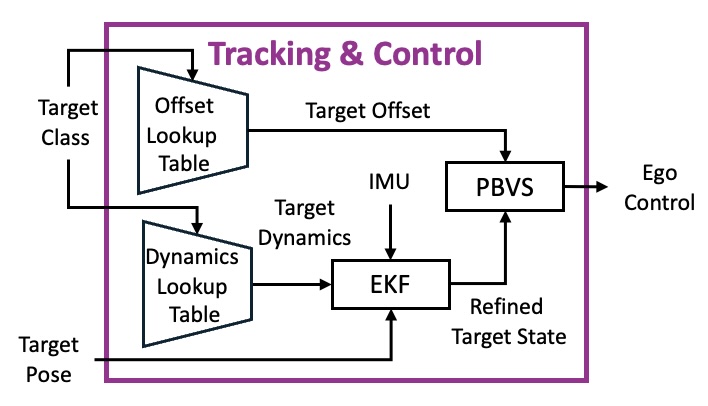}
    \caption{\small{
    \textbf{Physics-aware tracking and control:} the predicted target class is used to select class-specific tracking offsets and dynamics priors, and to fuse raw pose estimates with ego IMU in an EKF to obtain refined target state estimates. A pose-based visual servoing controller (PBVS) then commands the ego quadrotor for tracking.
    }}
    \label{fig:tracking-control-module}
\end{figure}

\begin{table*}[htbp]
\centering
\caption{\small{Quantitative perception performance across 3 objects in FalconGym 2.0 simulation and the real world. We report class accuracy, IoU, MTE and MAE across targets. Compared to geometric PnP and direct pose regression, our proposed unified multi-head architecture, staged training algorithm and FalconGym 2.0 reprojection (RP) consistency all contribute to better perception performance.}}
\label{tab:perception-result}
\scriptsize
\resizebox{\textwidth}{!}{
\begin{tabular}{lccc|cccc|cccc}\toprule
    & & & & \multicolumn{4}{|c|}{\textbf{FalconGym 2.0 (2 Environments)}} & \multicolumn{4}{c}{\textbf{Real World (2 Environments)}} \\\midrule
    Object & Method & Auto-Label? & Unified? & Class (\%)$\uparrow$ & IoU$\uparrow$ & MTE (\%)$\downarrow$ & MAE (rad)$\downarrow$ &  Class (\%)$\uparrow$ & IoU$\uparrow$ & MTE (\%)$\downarrow$ & MAE (rad)$\downarrow$ \\\midrule
    
    F1-tenth & PnP \cite{7368948} & \xmark & \xmark & - & - & 52\% & 0.88 & - & -\footnotemark & 82\% & 1.21 \\
    & NPE \cite{11247178} & \cmark & \xmark & - & - & 42\% & 0.54 & - & - & 52\% & 0.77 \\
    & Ours (no stage)  & \cmark & \cmark & 95\% & 0.74 & 40\% & 0.57 & 96\% & - & 57\% & 0.78 \\
    & Ours (no RP)  & \cmark & \cmark & 100\% & 0.81 & 41\% & 0.52 & 100\% & - & 55\% & 0.74 \\
    & \gb{Ours (full)} & \gb{\cmark} & \gb{\cmark} & \gb{100\%} & \gb{0.83} & \gb{24\%} & \gb{0.27}  & \gb{100\%} & \gb{-} & \gb{43\%} & \gb{0.38}\\
    \midrule

    Quadrotor & PnP \cite{7368948} & \xmark & \xmark & - & - & 71\% & 1.12 & - & - & 82\% & 0.82 \\
    & NPE \cite{11247178} & \cmark & \xmark & - & - & 54\% & 0.52 & - & - & 61\% & 0.51  \\
    & Ours (no stage) & \cmark & \cmark & 100\% & 0.72 & 56\% & 0.48 & 96\% & - & 53\% & 0.56  \\
    & Ours (no RP) & \cmark & \cmark & 100\% & 0.79 & 51\% & 0.55 & 96\% & - & 54\% & 0.53 \\
    & \gb{Ours (full)} & \gb{\cmark} & \gb{\cmark} & \gb{100\%} & \gb{0.79} & \gb{32\%} & \gb{0.31} & \gb{96\%} & \gb{-} & \gb{41\%} & \gb{0.37} \\
    \midrule

    Gate & PnP \cite{7368948} & \xmark & \xmark & - & - & 60\% & 0.76 & - & - & 72\% & 0.69 \\
    & NPE \cite{11247178} & \cmark & \xmark & - & - & 54\% & 0.74 & - & - & 68\% & 0.74   \\
    & Ours (no stage) & \cmark & \cmark & 96\% & 0.65 & 50\% & 0.71 & 98\% & - & 69\% & 0.69  \\
    & Ours (no RP) & \cmark & \cmark & 100\% & 0.72 & 52\% & 0.65 & 100\% & - & 71\% & 0.71 \\
    & \gb{Ours (full)} & \gb{\cmark} & \gb{\cmark} & \gb{100\%} & \gb{0.72} & \gb{41\%} & \gb{0.41} & \gb{100\%} & \gb{-} & \gb{52\%} & \gb{0.52}\\
    \midrule

\end{tabular}
}
\end{table*}

Given the target class and pose estimated by perception, we design a physics-aware tracking and control pipeline (Figure~\ref{fig:tracking-control-module}). 
Prior vision-based tracking work often relies on fiducial markers \cite{LiYangMitra, 9981921}, which are intrusive and impractical in many real settings.  
Other work relies on variations of Image-Based Visual Servoing (IBVS) using a 2D bounding-box center \cite{10610111}, but 2D boxes can be ambiguous under perspective projection and do not reliably capture target yaw, which is critical for stable tracking. 
We therefore track the full 6-DoF pose in the camera frame and inject class-specific motion priors to resolve target dynamics.

\begin{figure*}[htbp]
    \centering
    \includegraphics[width=0.9\linewidth]{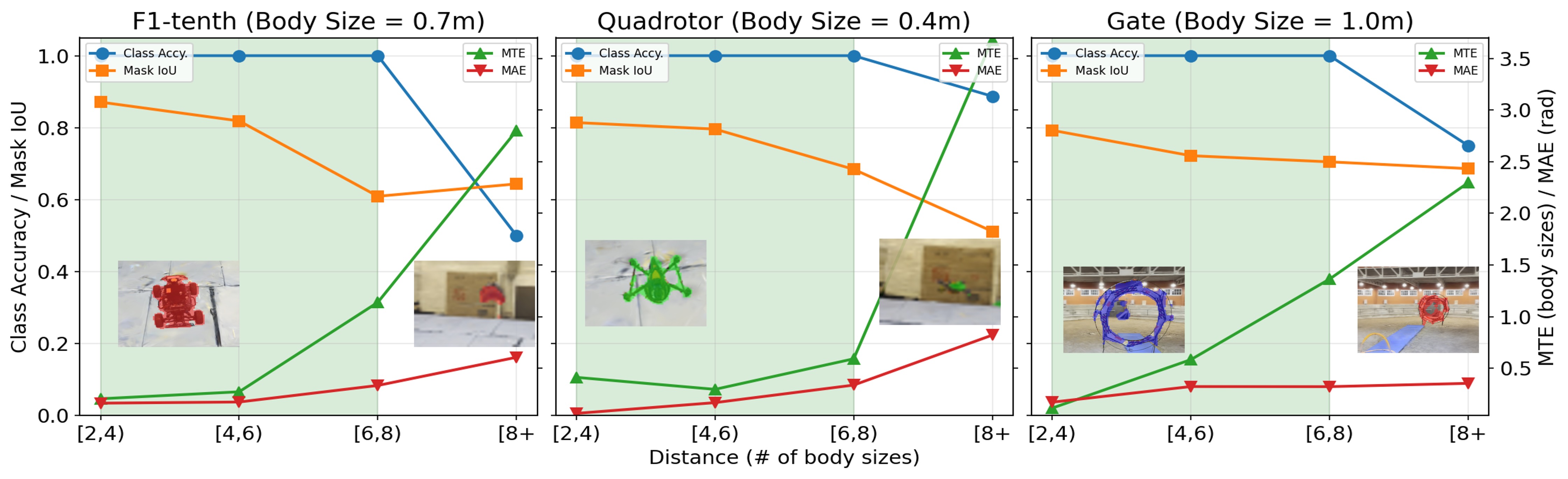}
    \caption{\small{
    \textbf{Operational Design Domain (ODD) of Our Perception Module:} We evaluate classification accuracy, mask IoU, normalized mean translation error (MTE, in body lengths), and mean angular error (MAE) as a function of target distance (in body lengths) for three target classes. The quantitative results indicate an ODD (annotated in green) of 2--6 body lengths for reliable pose estimation.
    }}
    \label{fig:ODD}
\end{figure*}

\paragraph{Desired Tracking Offset}
For each target class, we define a desired relative pose offset between the ego camera and the target to capture task-specific stand-off distance. 
For a gate, the offset centers the gate at a fixed forward distance (2 m); for vehicles, we keep a forward distance (2 m) with a vertical offset (1 m) for visibility and to reduce ground effects. 
All offsets lie within the perception module's operating design domain, as shown by Figure \ref{fig:ODD} and discussed by Section \ref{sec:ODD}, to ensure the target remains in view.

\paragraph{Dynamics-Aware State Estimation}
\label{sec:physics-aware-tracking}
We maintain the target state in the ego frame (coincident with the camera frame) and estimate it using an EKF that fuses raw pose measurements with class-dependent dynamics. 
The process model is selected from a dynamics lookup table (Figure \ref{fig:tracking-control-module}): Dubins car for F1-tenth, 12-state rigid-body dynamics for quadrotors \cite{Sabatino2015QuadrotorCM}, and a double-integrator model for gates. 
This EKF smooths noisy pose estimates and propagates the target state through brief measurement dropouts when the target is temporarily out of view.

\paragraph{Pose-based visual servoing (PBVS)}

PBVS takes the EKF-estimated target pose in the ego body frame and the desired relative offset from the lookup table, and outputs body-frame commands $(\mathbf{v}, \dot{\psi})$, where $\mathbf{v} \in \mathbb{R}^3$ is the commanded body-frame velocity and $\dot{\psi}$ is the commanded body-frame yaw rate. 
Let $\hat{\mathbf{p}}, \hat{\mathbf{v}} \in \mathbb{R}^3$ and $\hat{\psi}$ denote the EKF-estimated target position, velocity, and yaw in the ego body frame, and let $\mathbf{p}_d, \psi_d$ be the desired offset. 
We define body-frame errors
\begin{equation}
\mathbf{e}_p = \hat{\mathbf{p}} - \mathbf{p}_d, \quad \dot{\mathbf{e}}_p = \hat{\mathbf{v}}, \quad e_\psi = \mathrm{normalize}(\hat{\psi}-\psi_d),
\end{equation}
and apply a PD law in translation with a proportional term in yaw. 
Empirically, we set $K_p = [0.5, 0.2, 0.05], K_d = [0.1, 0.05, 0.01], k_\psi = 0.5$

\begin{equation}
\mathbf{v} = -K_p \odot \mathbf{e}_p - K_d \odot \dot{\mathbf{e}}_p, \quad \dot{\psi} = -k_\psi e_\psi.
\end{equation}

\footnotetext{Both baseline perception methods do not output class predictions or masks, and real-world ground-truth mask segmentation is unavailable without manual per-pixel annotation}

\section{Experiments}
In this section, we first evaluate the perception module in isolation across three target objects in both seen and unseen backgrounds, including sim-to-real comparisons (Section \ref{sec:perception-analysis}). 
We then evaluate end-to-end FalconTrack in closed-loop tracking over diverse trajectories and backgrounds, again reporting sim-to-real results (Section \ref{sec:tracking-control-analysis}).

\subsection{Perception Module Analysis}
\label{sec:perception-analysis}
We first report quantitative perception performance in simulation and the real world, then analyze the operational design domain (ODD).

\subsubsection{Metrics \& Baselines}
Our perception module outputs a target class label, a segmentation mask, and a 6-DoF pose estimate.
As shown in Table \ref{tab:perception-result}, we evaluate perception using four metrics: classification accuracy (\%), mask intersection-over-union (IoU), mean translational error (MTE), and mean angular error (MAE). 
To compare across targets of different sizes, we report MTE as a percentage of body size, while MAE is reported in radians.

We compare against two pose-estimation baselines: a traditional perspective-$n$-point (PnP) method \cite{7368948} and a neural pose estimator (NPE) \cite{11247178}. 
PnP typically requires manual keypoint labels to obtain 3D--2D correspondences. Given the large training set ($\sim$10k), we adapt our automated labeling pipeline (Figure \ref{fig:auto-label}) to automatically select 2D points from 3D Gaussians, while still manually selecting the Gaussians. 
Even so, PnP remains per-object and requires manual feature selection, so it is marked \xmark\, in ``Unified?'' and ``Auto-label?'' in Table \ref{tab:perception-result}.
The NPE baseline uses auto-labels from a NeRF-based simulator, but it is trained per object and per background; therefore, it is also marked \xmark\, in ``Unified?''.
Both baselines output only pose estimates; therefore, we leave classification and IoU blank in Table \ref{tab:perception-result}.
%
Our unified perception model uses auto-labels and generalizes across three target classes without per-target training. 
We include three variants for ablation: Ours (no stage) trains all heads jointly without staged training (Algorithm \ref{alg:perception-training}) or reprojection loss; Ours (no RP) uses staged training but disables the reprojection loss; and Ours (full) includes both.
For fair comparison, each method uses the full 10k perception dataset from Section \ref{sec:perception} with the same training budget; all three of our variants are trained for 250 epochs.

\subsubsection{Perception Evaluation}
We first evaluate five perception methods in FalconGym 2.0 across three target objects and two environments, using 200 unseen validation samples per object per background. 
We then evaluate on $\sim$200 real-world images per target per background collected by hand-holding the onboard camera. 
Real-world ground truth is obtained from a motion-capture system (millimeter-level accuracy) and transformed into the camera frame.
Results in Table \ref{tab:perception-result} show that PnP has the largest MAE and MTE across objects, especially for the quadrotor, likely due to imperfect manual feature selection and sensitivity to irregular geometry and partial occlusions. 
NPE performs better than PnP but still requires per-object, per-background training, which limits scalability to more objects or backgrounds.
All three of our variants generalize across objects and backgrounds and outperform PnP and NPE overall (thus marked \cmark\, in Table \ref{tab:perception-result}).
Staged training slightly improves classification and IoU in both simulation and the real world, while reprojection loss yields the largest gains in pose accuracy (MTE and MAE). 
Because pixel-level ground-truth masks are unavailable for real-world images without manual annotation, we report only class and pose metrics for real-world evaluation.
Qualitative results (Figure \ref{fig:perception-unknown-background}) further show consistent perception in unseen simulated and real environments with multiple distractors.
A sim-to-real gap is observed for all methods, as expected due to lighting and appearance differences between simulation and real environments.

\begin{table*}[htbp]
\centering
\caption{\small{Tracking performance across objects in FalconGym 2.0 and hardware. We report Success, Target-in-FOV, and ATE for baselines and ablations. Real quadrotor tracking quadrotor experiments are skipped due to safety constraints and single-quadrotor availability.}}
\label{tab:tracking-result}
\scriptsize
\resizebox{\textwidth}{!}{
\begin{tabular}{lcccc|ccc}\toprule
    & & \multicolumn{3}{c|}{FalconGym 2.0 (5 trajectories in 2 environments)} & \multicolumn{3}{c}{Real World (5 trajectories in 2 environments)} \\\midrule
    Object & Method & Success(\%) $\uparrow$ & Target in FOV(\%) $\uparrow$ & ATE (m) $\downarrow$ & Success(\%) $\uparrow$ & Target in FOV(\%) $\uparrow$ & ATE (m) $\downarrow$ \\\midrule
    
    F1-tenth & \underline{State-based} & \underline{100\%} & \underline{100\%} & \underline{0.37} & \underline{100\%} & \underline{100\%} & \underline{0.45} \\
    & Mask-centered IBVS \cite{10610111} & 80\% & 84\% & 0.77 & 60\% & 68\% & 1.23  \\
    & Ours (w/o physics) & 100\% & 100\% & 0.57 & 100\% & 95\% & 0.92 \\
    & \gb{Ours (w/ physics)} & \gb{100\%} & \gb{100\%}  & \gb{0.50} & \gb{100\%} & \gb{99\%} & \gb{0.71}\\
    \midrule

    Quadrotor & \underline{State-based} & \underline{100\%} & \underline{100\%} & \underline{0.45} & - \footnotemark & - & - \\
    & Mask-centered IBVS \cite{10610111} & 60\% & 72\% & 0.94 & - & - & - \\
    & Ours (w/o physics) & 80\% & 95\% & 0.72 & - & - & - \\
    & \gb{Ours (w/ physics)} & \gb{100\%} & \gb{100\%} & \gb{0.66} & - & - & - \\
    \midrule
    
    Gate & State-based & \underline{100\%} & \underline{100\%} & \underline{0.11} & \underline{100\%} & \underline{100\%} & \underline{0.33} \\
    & Mask-centered IBVS \cite{10610111} & 100\% & 100\% & 0.19 & 100\% & 100\% & 0.47 \\
    & Ours (w/o physics) & 100\% & 100\% & 0.27 & 100\% & 100\% & 0.67 \\
    & \gb{Ours (w/ physics)} & \gb{100\%} & \gb{100\%} & \gb{0.18} & \gb{100\%} & \gb{100\%} & \gb{0.43}\\
    \midrule
    
\end{tabular}
}
\end{table*}

\subsubsection{Perception ODD}
\label{sec:ODD}
We further analyze the operational design domain (ODD) of the perception module. 
We sample 200 images at different target distances (2--4, 4--6, 6--8, and $>8$ body lengths) with random poses and background in FalconGym 2.0. 
For this ODD analysis, we use the ``Ours (full)'' method.
As shown in Figure \ref{fig:ODD}, for each target, perception performance across all four metrics degrades with distance as the target occupies fewer pixels. 
Based on this analysis, we define the reliable ODD as 2--6 body lengths. 
In hardware experiments, we enforce a geo-fence safety check and abort trials if the ego vehicle exceeds 8 body lengths from the target.
We set the geo-fence threshold to 8 body lengths because our physics-aware tracking controller can sometimes recover from brief one-frame perception failures.

\subsection{Tracking \& Control Evaluation} 
\label{sec:tracking-control-analysis}

We evaluate the end-to-end FalconTrack stack in FalconGym 2.0 and on hardware. We first describe the platform and arenas, then define metrics and baselines, and finally present a sim-to-real analysis.

\subsubsection{Hardware Quadrotor Setup}
We use a customized quadrotor equipped with an NVIDIA Jetson Orin companion computer (Ubuntu 20.04, PyTorch) and a Betaflight flight controller. The RGB camera is an ArduCam streaming at 800$\times$600 at 100\,Hz.
To match training and improve inference speed, we downscale images to 224$\times$224. The Jetson runs the 15M-parameter perception model at $\sim$30\,Hz during offline tests. 
During real flights with motors spinning, perception inference, tracking control, and data logging, the closed-loop stack runs at $\sim$25\,Hz.

\subsubsection{Hardware Arena Setup}
For safety, and because only one quadrotor is available, we run hardware experiments for quadrotor tracking of F1-tenth and gate targets. Quadrotor-tracking-quadrotor is evaluated only in FalconGym 2.0.
For hardware gate-tracking experiments, we attach a string to the gate stand and manually pull it sideways for 10 meters. 
We run five trials across two environments with slightly different pulling speeds and log the gate trajectory with motion capture. 
This logging lets us replay the exact trajectory in FalconGym 2.0 for accurate sim-to-real comparison. 
We limit the ego quadrotor speed to 0.75\,m/s because the flight path moves toward the spectator area.
For F1-tenth tracking, we record five trajectories (two straight trajectories with different speeds, one left turn, one right turn, and one S-shape) across two environments per tracking method. 
A skilled driver operates the F1-tenth target along a predefined route at 1--2\,m/s. 
As with gate tracking, we also log the F1-tenth pose via motion capture and replay it in FalconGym 2.0. 
We cap the ego quadrotor speed at 2.5\,m/s, slightly above the F1-tenth maximum speed (2.25\,m/s) in our arena.
During both real closed-loop flights (gate and F1-tenth tracking), motion-capture measurements are used only for offline logging and evaluation and are not provided to the ego quadrotor. 

\subsubsection{Metrics \& Baselines}
As shown in Table \ref{tab:tracking-result}, we measure tracking performance with three metrics: Success (the ego quadrotor completes the run without operator intervention or geo-fence abort), Target in FOV (\% of frames where the target is visible), and ATE (average tracking error between the ego quadrotor and the desired relative offset). 
All trajectory visualization and ATE computation use motion-capture logs offline; no motion-capture target ground truth is available to the ego quadrotor during closed-loop tracking.
We compare against two tracking baselines and two variants of our method. Because we track a single target per run, the evaluation focuses on tracking quality and robustness.
The state-based baseline uses motion-capture ground-truth target pose, rather than the perception estimate in Figure \ref{fig:tracking-control-module}, to drive PBVS. 
The mask-centered IBVS baseline \cite{10610111} uses a PD controller to keep the ego quadrotor aligned with the center of the detected 2D bounding box. 
Ours (w/o physics) feeds raw perception pose directly to PBVS, while Ours (w/ physics) uses the full pipeline in Figure \ref{fig:tracking-control-module}, where perception is first refined by an EKF with dynamics priors before PBVS.

\subsubsection{Tracking Evaluation}
For all three objects, the state-based baseline achieves the best performance, as expected when ground-truth target pose is available. 
The 100\% Target-in-FOV score of the state-based baseline also indicates that continuously keeping the target in view is feasible for these trajectories (i.e., the benchmark is not dominated by dynamically infeasible paths).
The mask-centered IBVS baseline performs well in most cases, but struggles when the target suddenly leaves the field of view; this is most evident in the F1-tenth S-shape trajectory, where rapid turns induce large yaw changes that IBVS cannot recover from.
Ours (w/o physics) occasionally loses the target for several frames when tracking the F1-tenth vehicle, but the 6-DoF pose estimate (including yaw) allows recovery.
Adding dynamics-aware estimation (Ours w/ physics) further improves smoothness and tracking accuracy across both simulation and real hardware, especially for out-of-view recovery enabled by the physics-aware tracking approach in Section \ref{sec:tracking-controller}.

Although we have been able to track longer and more aggressive trajectories (e.g., extended spatial S-shapes) in FalconGym 2.0, we do not test those in hardware due to current safety constraints (missing safety net and inadequate safety padding).
For fair sim-to-real comparison, we replay measured hardware trajectories in FalconGym 2.0 by loading the object GSplat and driving it with the same motion-capture trajectories, using the coordinate transforms and Edit API from \cite{MiaoEtAl:ICRA26}.
Overall, sim-to-real trends are consistent: success and target-in-FOV remain high, while ATE increases in real hardware due to unmodeled disturbances (lighting changes, motion blur, and aerodynamic effects).

\footnotetext{Real-world quadrotor tracking quadrotor experiments are skipped due to safety constraints and single-quadrotor availability.}
\section{Conclusion \& Discussion}

We present FalconTrack, a unified perception-and-control framework for vision-based aerial tracking of dynamically moving targets using an onboard RGB camera.
FalconTrack combines a multi-head perception model trained with automatically generated labels in FalconGym 2.0 with a physics-aware tracking controller that fuses perception with class-specific dynamics priors.
Experiments in both simulation and hardware demonstrate strong perception and tracking performance relative to baselines, and show successful zero-shot sim-to-real transfer.

In future work, we plan to address three limitations.
First, we assume a single target per run and do not yet handle target switching or multi-object tracking.
Second, the perception pipeline currently assumes known object instances and requires a short video for GSplat reconstruction, which limits open-world deployment to unknown objects.
Third, hardware validation does not yet cover very fast targets (e.g., $>2$\,m/s), where motion blur is more pronounced, and real-world quadrotor-tracking-quadrotor experiments remain to be completed once stronger arena safety measures are in place.

\bibliographystyle{IEEEtran}
\bibliography{reference}










\end{document}